\documentclass[sigconf]{acmart}

\usepackage{booktabs} 
\usepackage{todonotes}
\usepackage{amssymb}
\usepackage{pifont}
\newcommand{\xmark}{\ding{54}}%

\setcopyright{none}
\setcopyright{rightsretained}

\begin{document}
\title{On the Flip Side: Identifying Counterexamples in Visual Question Answering}

\author{Gabriel Grand}
\affiliation{%
  \institution{Harvard University}
}
\email{ggrand@college.harvard.edu}

\author{Aron Szanto}
\affiliation{%
  \institution{Harvard University}
}
\email{aszanto@college.harvard.edu}

\author{Yoon Kim}
\affiliation{%
  \institution{Harvard University}
}
\email{yoonkim@seas.harvard.edu}

\author{Alexander Rush}
\affiliation{%
  \institution{Harvard University}
}
\email{srush@seas.harvard.edu}



\begin{abstract}
Visual question answering (VQA) models respond to open-ended natural language questions about images. While VQA is an increasingly popular area of research, it is unclear to what extent current VQA architectures learn key semantic distinctions between visually-similar images. To investigate this question, we explore a reformulation of the VQA task that challenges models to identify counterexamples: images that result in a different answer to the original question. We introduce two methods for evaluating existing VQA models against a supervised counterexample prediction task, VQA-CX. While our models surpass existing benchmarks on VQA-CX, we find that the multimodal representations learned by an existing state-of-the-art VQA model do not meaningfully contribute to performance on this task. These results call into question the assumption that successful performance on the VQA benchmark is indicative of general visual-semantic reasoning abilities.
\end{abstract}

%
%



\maketitle

\section{Introduction}
\label{sec:introduction}

Visual question answering (VQA) is an increasingly popular research domain that unites two traditionally disparate machine learning subfields: natural language processing and computer vision. The goal of VQA is to generate a natural language answer to a question about an image. While a number of existing approaches perform well on VQA \citep{Gupta17survey,wu2017survey}, it is unclear to what extent current models are capable of making semantic distinctions between visually-similar images.

\begin{figure}[ht]
\vspace{0.25cm}
\begin{center}
\centerline{\includegraphics[width=\columnwidth]{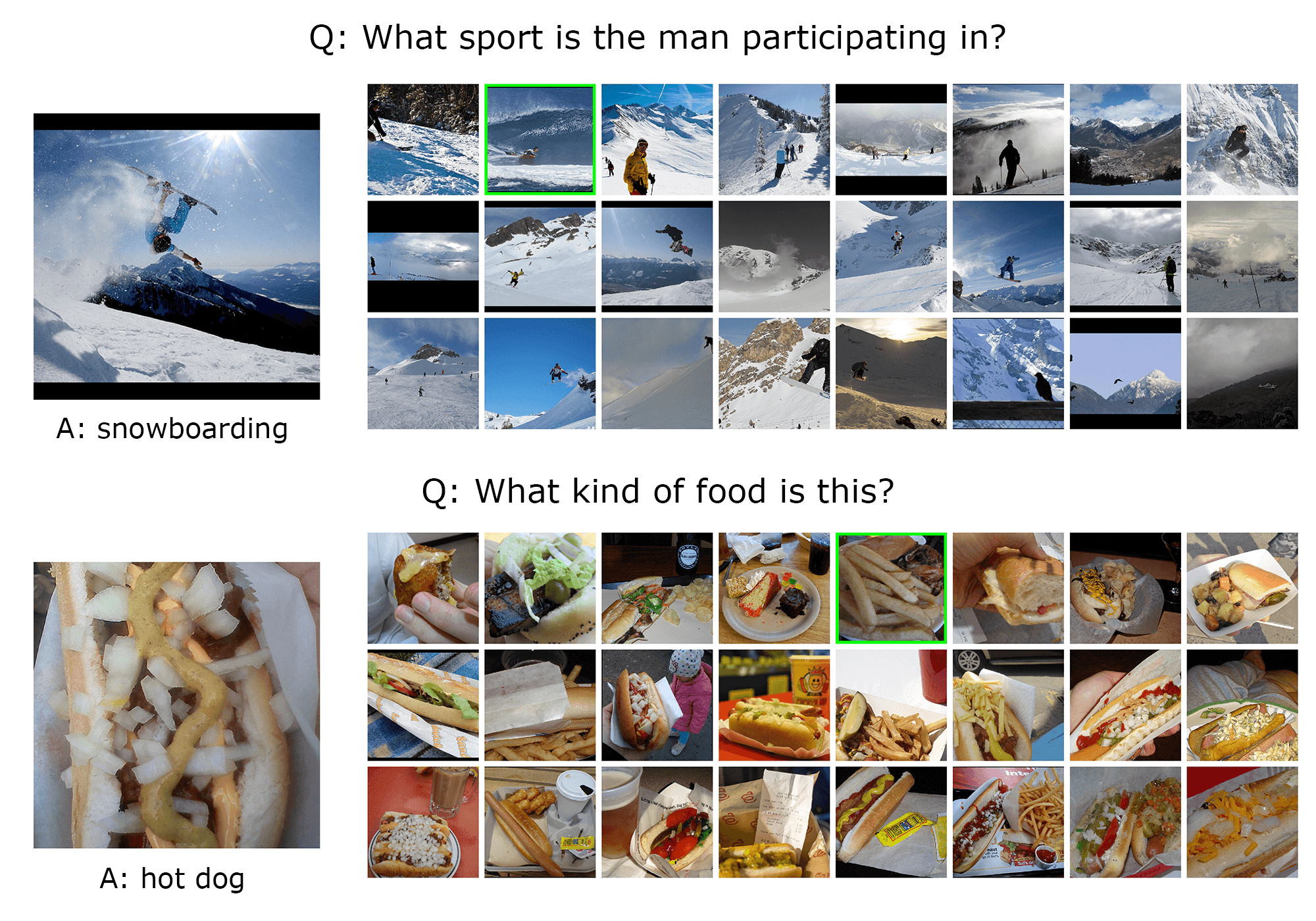}}
\vspace{-0.25cm}
\caption{Counterexample task. The goal is to identify a counterexample (green border) that results in a different answer to the question from a set of 24 visually-similar images.}
\label{fig:cx_task}
\end{center}
\vspace{-0.70cm}
\end{figure}

In this work, we explore a reformulation of the VQA task that more directly evaluates a model's capacity to reason about the underlying concepts encoded in images. Under the standard VQA task, given the question ``What color is the fire hydrant?'' and an image of a street scene, a model might answer ``red.'' Under the alternative task, the model must produce a counterexample; e.g., an image of a fire hydrant that is not red. Successful performance on the visual counterexample prediction task (abbreviated VQA-CX) requires reasoning about how subtle visual differences between images affect the high-level semantics of a scene.

The VQA-CX task was originally proposed in \citet{VQA2} as a useful explanation modality for VQA models. However, despite its applicability as a powerful tool for model introspection, this idea has remained largely under-explored by the research community. To our knowledge, this work represents the first follow-up attempt to operationalize the VQA-CX paradigm originally proposed by \citet{VQA2}.

We introduce two plug-and-play approaches for evaluating the performance of existing, pretrained VQA models on VQA-CX. The first method is an unsupervised model that requires no training and works out-of-box with a pretrained VQA model. The second method is a supervised neural model that can be used with or without a pretrained VQA model. The unsupervised model outperforms the baselines proposed in \citet{VQA2}. Meanwhile, the supervised model outperforms all existing unsupervised and supervised methods for counterexample prediction.

Crucially, while we use a state-of-the-art VQA model to facilitate counterexample prediction, we find that our methods perform almost as well without receiving any information from this model. In other words, the multimodal representation learned by the VQA model contributes only marginally (approximately 2\%) to performance on VQA-CX. These results challenge the assumption that successful performance on VQA is indicative of more general visual-semantic reasoning abilities.

\section{Background}

Contemporary research interest in VQA began with the release of DAQUAR, the DAtaset for QUestion Answering
on Real-world images \citep{malinowski2014daquar}. Since then, at least five other major VQA benchmarks have been proposed. These include COCO-VQA \citep{ren2015cocovqa}, FM-IQA \citep{gao2015fm-iqa}, VisualGenome \citep{krishna2017visualgenome}, Visual7w \citep{zhu2016visual7w}, and VQA \citep{VQA1,VQA2}. With the exception of DAQUAR, all of the datasets use include images from the Common Objects in Context (COCO) dataset \citep{lin2014coco}, which contains 330K images.

The VQA dataset was first introduced in \citet{VQA1} as more free-form, open-ended VQA benchmark. Previous datasets placed constraints on the kinds of questions authored by human annotators (e.g., Visual7w, VisualGenome), or relied on image captioning models to generate questions (e.g., COCO-VQA). In contrast, the crowdsourcing method employed by \citet{VQA1} was designed generate a more diverse range of question types requiring both visual reasoning and common knowledge. However, owing in part to the lack of constraints on question generation, the original VQA dataset contains several conspicuous biases. For instance, for questions beginning with the phrase, ``What sport is...'', the correct answer is ``tennis'' 41\% of the time. Additionally, question generation was impacted by a visual priming bias \citep{zhang2016yin}, which selected for questions with affirmative answers. For instance, for questions beginning with ``Do you see a...,'' the correct answer is ``yes'' 87\% of the time. Models that exploit these biases can achieve high accuracy on VQA without understanding the content of the accompanying images \citep{VQA2}.

In an effort to balance the VQA dataset, \citet{VQA2} introduced VQA 2.0, which is built on pairs of visually-similar images that result in different answers to the same question. Specifically, for each image $I$ in the original dataset, \citet{VQA2} determined the 24 nearest neighbor images $I_\mathrm{NN} = \{I'_1, \ldots, I'_{24}\}$ using convolutional image features $V$ derived from VGGNet \citep{simonyan2014VGG}. For each image/question/answer pair $(I, Q, A)$ in the original VQA dataset, crowd workers were asked to select a complementary image $I^* \in I_\mathrm{NN}$ that produced a different answer $A^*$ for the same $Q$. The most commonly selected $I^*$ was then included as a new example $(I^*, Q, A^*)$ in VQA 2.0, resulting in a dataset that is roughly double the size of the original. In addition to reducing language biases in the data, the pair-based composition of VQA 2.0 provides a convenient approach for supervised training and evaluation of counterexample prediction models.

\section{Approach}

We treat VQA-CX as a supervised learning problem, which can be formalized as follows. For each image, question, and answer $(I, Q, A)$ in the original VQA task, the model is presented with the the $K=24$ nearest neighbor images $I_\mathrm{NN} = \{I'_1, \ldots, I'_{K} \}$ of the original image. The model assigns scores $\mathcal{S} = S(I'_1), \ldots, S(I'_{K})$ to each candidate counterexample. The crowd-selected counterexample $I^* \in I_\mathrm{NN}$ serves as ground truth. For notational clarity, we distinguish between raw images $I$ and convolutional image features $V$. Additionally, we use prime notation ($I', V', A'$) to denote candidate counterexamples, asterisk notation to denote the ground truth counterexample ($I^*, V^*, A^*$), and no superscript when referring to the original example $(I, V, A)$. We do not use any superscripts for $Q$, since the question is the same in all cases.

Both of our VQA-CX models use an existing VQA model as a submodule. While there exist many diverse solutions for VQA \citep{wu2017survey}, we mostly treat the VQA model as a black box that can be expressed as a function of its inputs. We make only two assumptions about the architecture. First, we assume the model outputs a distribution $P(\mathcal{A}|I, Q)$ over a discrete number of answer classes (where $|\mathcal{A}|$ is a hyperparameter).\footnote{While most models treat VQA as a discrete classification task, some adopt a generative approach (e.g., \citet{wu2016ask,zhu2015building,wang2017fvqa}), which is not compatible with our methods.} Second, we assume the model internally combines its inputs into some multimodal representation $Z$, which we can access. (Note that this second assumption, which violates the black box principle, is only used optionally in the NeuralCX model.) We therefore treat a VQA model as a function $\text{VQA}(I, Q) = P(\mathcal{A}|I, Q), Z$.

In order to establish a basis for comparison with \citet{VQA2}, we began by reproducing their baselines, described in the following section. We then developed two architectures for VQA-CX. Both models can be used in conjunction with any VQA model that meets the above two criteria. The first architecture, which we call the Embedding Model, compares the semantic similarity between candidate answers in an embedding space, weighing different answers by $P(\mathcal{A}|I, Q)$. Since the Embedding Model relies solely on a pretrained VQA model and a pretrained answer embedding, it is fully unsupervised and requires no training. The second architecture is a straightforward multilayer perceptron that takes as input features related to $I$, $I'$, $Q$, and $A$, including the outputs of a VQA model, and returns a score $S(I')$. This NeuralCX model is trained in a pairwise fashion using standard supervised learning methods.

\section{Models}

\subsection{Prior Work}

To our knowledge, the only previous work on VQA-CX was carried out by the authors of the VQA 2.0 dataset. \citet{VQA2} present a two-headed model that simultaneously answers questions and predicts counterexamples. The model consists of three components:

\smallskip

\noindent\textbf{Shared base: }Produces a multimodal embedding of the question and image via pointwise multiplication, as in \citep{lu2015deeper}.
\[Z = \text{CNN}(I) \cdot \text{LSTM}(Q)\]
During a single inference step, a total of $K + 1$ images (the original image and its KNNs) are passed through is component.

\smallskip

\noindent\textbf{Answering head: }Predicts a probability distribution over answer classes.
\[P(\mathcal{A}|Z) = \sigma(W_{out}Z + b_{out})\]
Only the $Z$ corresponding to the original image is used in the answering head.

\smallskip

\noindent\textbf{Explaining head: }Predicts counterexample scores for each of $K$ nearest neighbor images.
\[S(I'_i) = (W_{zd} Z_i + b_{zd}) \cdot (W_{ad} A + b_{ad})\]

This component can be seen as computing vector alignment between a candidate counterexample and the ground truth answer. To allow for the dot product computation, $Z_i$ and $A$ are both projected into a common embedding space of dimensionality $d$. Note that in the final layer of the network, all $K$ scores $\mathcal{S} = S(I'_1), \ldots, S(I'_{K})$ are passed through a $K \times K$ fully-connected layer, presumably to allow the model to learn the distribution over the rank of $I^*$ within $I_\mathrm{NN}$.

The two-headed model is trained on a joint loss that combines supervision signals from both heads. 

\[\mathcal{L(S)} = -\log P(A | I, Q) + \lambda \sum_{I'_i \neq I^*} \max(0, M - (S(I^* - I'_{i})))\]

The answer loss is simply the cross entropy loss induced by the ground truth answer $A \in \mathcal{A}$. Meanwhile, the explanation loss is a pairwise hinge ranking loss \citep{chopra2005similarity}, which encourages the model to assign the ground-truth counterexample $I^*$ a higher score than the other candidates.

\subsection{Baselines}

In addition to their counterexample model, \citet{VQA2} introduce three key baselines for VQA-CX:

\begin{itemize}
\item \textbf{Random Baseline:} Rank $I_\mathrm{NN}$ randomly.
\item \textbf{Distance Baseline:} Rank $I_\mathrm{NN}$ by L2 distance from $I$. Closer images are assigned higher scores.
\item \textbf{Hard Negative Mining:} For each $I'_i \in I_\mathrm{NN}$, determine the probability of the original answer $P(A)_i = \text{VQA}(I'_i, Q)$ using a pretrained VQA model. Rank the $I'_i$ according to \textit{negative} probability $-P(A)_i$. In other words, choose counterexamples for which the VQA model assigns a low probability to the original answer.
\end{itemize}

\subsection{Unsupervised Embedding Model}

Successful performance on VQA-CX requires a nuanced treatment of the semantic relationship between answers. While the counterexample answer $A^*$ is distinct from the original answer $A$, the two are often close neighbors in semantic space. For example, for the question-answer pair ($Q=$ ``What animal is in the tree?''; $A=$``cat''), the counterexample answer is more likely to be ``dog'' than ``meatball,'' even though the semantic distance between ``cat'' and ``meatball'' is greater. Ideally, a VQA-CX model should take into account this linguistic prior.

The Embedding Model counterbalances the goal of identifying a semantically-similar counterexample answer with the necessity that the answer not be identical to the original. The model uses answer-class predictions $P(\mathcal{A} | I', Q)$ from a pretrained VQA model, and answer embeddings $W_\mathcal{A}$ from a pretrained Skip-Thoughts model \citep{skipthoughts} to assign a score to each nearest neighbor image:
\begin{multline}
S(I'_i) = \lambda \sum_{\substack{a \in \mathcal{A};\\ a \neq A}} \text{cossim} \left(a, A\right) P(a|I', Q) - (1-\lambda)\log P(A|I', Q)
\end{multline}

The term to the left of the subtraction encourages the model to select counterexamples that produce answers similar to the original. Meanwhile, the term to the right discourages the model from selecting the exact same answer as the original. The $\lambda$ hyperparameter, chosen empirically, determines the relative weight of these terms.

\subsection{Supervised NeuralCX Model}

NeuralCX is a fully-connected network that takes as input 10 features derived from $I$, $I'$, $Q$, and $A$. Some of these features, such as $V$, $Q$, and $A$, are representations of the original image, question, and answer. Others, such as $Z$ and $P(\mathcal{A'})$, are computed by a VQA model. Table \ref{table:features} summarizes the input features.

\begin{figure}[ht]
\begin{center}
\centerline{\includegraphics[width=\columnwidth]{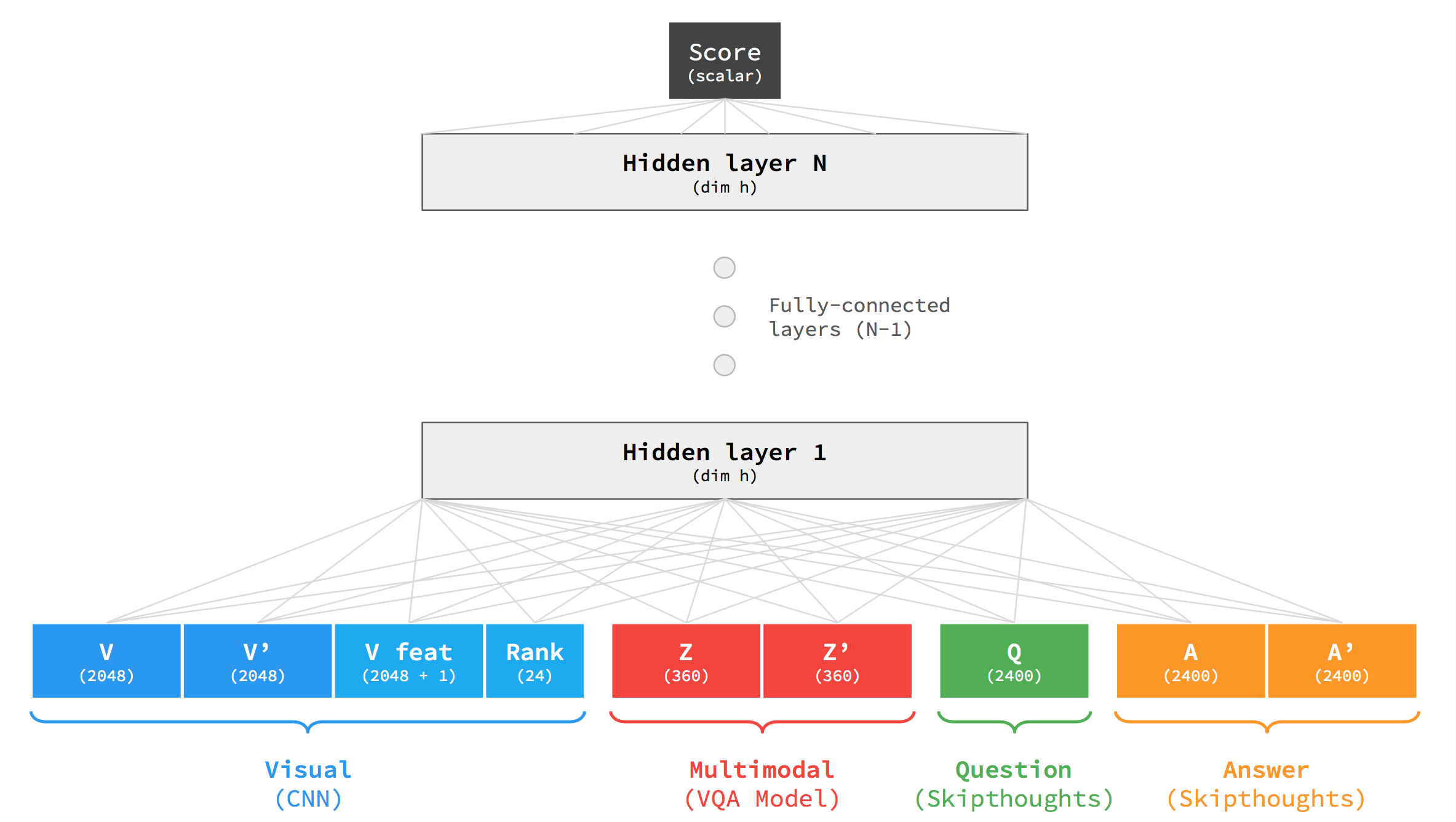}}
\vspace{-0.15cm}
\caption{Diagram of NeuralCX architecture. The model is a fully-connected neural network that takes visual, question, answer, and multimodal features as input and produces a score indicating the relevance of $I'$ as a counterexample.}
\label{fig:neuralcx}
\end{center}
\vspace{-0.15cm}
\end{figure}

All features are concatenated into a single input vector and passed through a series of hidden layers, where the size $h$ and number $N$ of layers are hyperparameters. All layers share the same $h$ and use ReLU activation. The output of the last hidden layer is projected to an unnormalized scalar score $S(I')$. Fig. \ref{fig:neuralcx} depicts the NeuralCX architecture.

A single training iteration for NeuralCX consists of $K$ forward passes of the network to produce a score for each candidate $I'_i \in I_\mathrm{NN}$. We compute the cross-entropy loss for the ground truth $I^*$, and optimize the parameters of the network via backpropagation.

\begin{table}[ht]
\vspace{-0.5cm}
\begin{center}
\begin{sc}
\begin{tabular}{@{}llll@{}}
\toprule
Feature & Definition & Size & Origin \\ \midrule \midrule
$V$ & $\text{CNN}(I)$ & $2048$ & CNN \\
$V'$ & $\text{CNN}(I')$ & $2048$ & CNN \\
$V_M$ & $V \odot V'$ & $2048$ & Computed \\
$V_D$ & $||V' - V||$ & $1$ & Computed \\
$\text{Rank}$ & $\text{onehot}(i)$ & $24$ & Computed \\ \midrule
$Q$ & $\text{LSTM}(Q)$ & $2400$ & LSTM \\
$A$ & $W_\mathcal{A} A$ & $2400$ & $A_\text{emb}$ \\
$A'$ & $(W_\mathcal{A})^T P(\mathcal{A'})$ & $2400$ & $A_\text{emb}$, VQA \\ \midrule
$Z$ & $\text{VQA}(I, Q)$ & $360$ & VQA \\
$Z'$ & $\text{VQA}(I', Q)$ & $360$ & VQA \\ \bottomrule
\end{tabular}
\end{sc}
\end{center}
\vspace{0.25cm}
\caption{Full set of features input to NeuralCX model.}
\label{table:features}
\vspace{-0.5cm}
\end{table}

\section{Methods}

Our VQA-CX dataset consists of 211K training examples and 118K test examples, of which 10K were reserved as a validation set. A single example in our dataset consists of the original VQA 2.0 example $(I, Q, A)$, and the 24 nearest neighbor images $I_\mathrm{NN}$, which contain the ground truth counterexample $I^* \in I_\mathrm{NN}$ and its corresponding answer $A^*$. 

Our train and test data are, by necessity, proper subsets of the VQA 2.0 training and validation datasets, respectively. To construct our trainset, we first identified the examples for which the image $I$ had a corresponding $I^*$. Approximately 22\% of the images in VQA 2.0 do not have a labeled complement. \footnote{These images correspond to instances in which crowd workers indicated that it was not possible to select a counterexample from among $I_\mathrm{NN}$ \citep{VQA2}.} Next, we filtered out examples for which $I^*$ did not appear in $I_\mathrm{NN}$. Since we used the nearest neighbors data provided by \citet{VQA2}, $I^*$ should theoretically always appear in $I_\mathrm{NN}$. However, because the KNN relation is not symmetric (i.e., $I^1 \in I^2_\mathrm{NN} \nRightarrow I^2 \in I^1_\mathrm{NN}$), we found that in certain cases, $I^* \notin I_\mathrm{NN}$. After filtering, we were left with $211,626 / 433,757$ train examples and $118,499 / 214,354$ validation examples. Note that while \citet{VQA2} collected labeled counterexamples for the VQA 2.0 dataset, this data is not public. As a result, we did not make use of the VQA 2.0 test set, instead testing on the VQA 2.0 validation set. 

We implemented our models and experiments in Pytorch \citep{paszke2017pytorch}. For all experiments involving VQA models, we used MUTAN \citep{MUTAN}, a state-of-the-art VQA model that uses Tucker decomposition to parametrize a bilinear interaction between $Q$ and $V$. We pretrained MUTAN separately on VQA 2.0 for 100 epochs with early stopping to a peak test accuracy of 47.70. Unfortunately, because we needed to train the model on only the VQA 2.0 training set (and not the validation set), this accuracy is considerably lower than the 58.16 single-model accuracy obtained by \citet{MUTAN}. Additionally, since the VQA-CX task requires us to load all 24 $V_{\mathrm{NN}}$ features into memory simultaneously, we opted use a no-attention variant of MUTAN that is more space-efficient, but lower-performing. We used a pretrained ResNet-152 model \citep{resnet152} to precompute visual features for all images, and a pretrained Skip-Thoughts model \citep{skipthoughts} to compute question and answer embeddings. We also utilized framework code from the vqa.pytorch Github repository.\footnote{https://github.com/Cadene/vqa.pytorch}

For all experiments with the NeuralCX model, we trained for a maximum of 20 epochs with early stopping. We optimize the model parameters with standard stochastic gradient descent methods, using the Pytorch library implementation of Adam \citep{DBLP:journals/corr/KingmaB14} with learning rate $0.0001$ and batch size $64$. We also employed dropout regularization ($p=0.25$) between hidden layers \citep{srivastava2014dropout}. 

We experimented with different numbers of hidden layers $N = 1, 2, 3$ and hidden units $h = 256, 512, 1024$, but found that larger architectures resulted in substantial training time increases with negligible performance gains. We therefore use a moderate-sized architecture of $N=2, h=512$ for all reported results. This model takes about 35 minutes to train to peak performance on a single Tesla K80 GPU.

We evaluate the performance of our models and baselines with recall@$k$, which measures the percentage of the ground truth counterexamples that the model ranks in the top $k$ out of the $24$ candidate counterexamples. Results on the test set for the NeuralCX Model, Embedding Model, and baseline models are reported in Table \ref{table:results}. To better understand the relative importance of the different inputs to the NeuralCX model, we selectively ablated different features by replacing them with noise vectors drawn randomly from a uniform distribution. We chose to randomize inputs, rather than remove them entirely, so as to keep the model architecture constant across experiments. In each ablation experiment, the model was fully retrained. Results from these experiments are reported in Table \ref{table:ablation}.

\section{Results}

We began by reimplementing the baselines presented by \citet{VQA2} and comparing our results with theirs. As expected, the Random Baseline performed approximately at chance (recall@5 $\approx \frac{5}{24}$ or $0.2083$). Our Distance Baseline was comparable with, but slightly higher than, the result reported by \citeauthor{VQA2}. This discrepancy indicates that it is possible that the distribution over the rank of the ground-truth counterexample is more skewed in our dataset than in the one used by \citeauthor{VQA2}. Notably, in both cases, the strategy of ranking counterexample images based on distance in feature space is more than two times better than chance, and serves as a strong baseline.

As in \citet{VQA2}, we found Hard Negative Mining to be a relatively under-performing approach. Since we used a different VQA model from \citeauthor{VQA2}, our results on this baseline are not directly comparable. Nevertheless, in both cases, Hard Negative Mining performed only marginally above chance. To isolate the impact of the VQA model, we computed the Hard Negative Mining baseline using an untrained (randomly initialized) VQA model. After this change, the performance dropped to random.

\begin{table*}[ht]
\begin{center}
\begin{sc}
\begin{tabular}{@{}lllll@{}}
\toprule
 &  & \multicolumn{2}{c}{Our Results} & \multicolumn{1}{c}{Goyal et al. (2016)} \\
\multicolumn{1}{l}{CX Model} & VQA Model & \multicolumn{1}{l}{Recall@1} & \multicolumn{1}{l}{Recall@5} & \multicolumn{1}{l}{Recall@5} \\ \midrule
Random Baseline & - & $4.20$ & $20.85$ & $20.79$ \\
Hard Negative Mining & untrained & $4.06$ & $20.73$ & - \\
Hard Negative Mining & pretrained & $4.34$ & $22.06$ & $21.65$ \\
Embedding Model & untrained & $4.20$ & $21.02$ & - \\
Embedding Model & pretrained & $7.77$ & $30.26$ & - \\
Distance Baseline & - & $11.51$ & $44.48$ & $42.84$ \\ \midrule
Two-headed CX (Goyal et al.) & trainable & - & - & $43.39$ \\
NeuralCX & untrained & $16.30$ & $52.48$ & - \\
NeuralCX & pretrained & $18.27$ & $54.87$ & - \\
\textbf{NeuralCX} & \textbf{trainable} & $\mathbf{18.47}$ & $\mathbf{55.14}$ & - \\ \bottomrule
\end{tabular}
\end{sc}
\end{center}
\vspace{0.25cm}
\caption{Results of VQA-CX models and baselines. Where applicable, we compare our results with those reported in \citet{VQA2}. The midline separates models that were evaluated without training (above) with those that were trained on the VQA-CX dataset (below). Untrained denotes that the VQA model parameters were randomly-initialized and immutable. Pretrained denotes parameters that were learned on the VQA task and then made immutable. Trainable denotes parameters that were first learned on VQA, and then fine-tuned on VQA-CX.}
\label{table:results}
\vspace{-0.5cm}
\end{table*}

\begin{table}[ht]
\begin{center}
\begin{small}
\begin{sc}
\begin{tabular}{@{}cc|ccccccc@{}}
\toprule
\multicolumn{2}{c|}{Performance} & \multicolumn{7}{c}{Ablated Features} \\
R@5 & R@1 & $V$ & $V_{\text{M}}$ & $V_{\text{D}}$ & Rank & $Q$ & $A$ & $Z$ \\ \midrule
$43.05$ & $12.33$ & \xmark & \xmark & \xmark & \xmark &  &  &  \\
$44.48$ & $11.42$ & \xmark &  &  &  &  &  &  \\
$44.48$ & $11.51$ &  & \xmark & \xmark & \xmark &  &  &  \\
$44.48$ & $11.52$ & \xmark & \xmark & \xmark &  & \xmark & \xmark & \xmark \\
$44.55$ & $13.17$ &  &  &  & \xmark &  &  &  \\
$47.09$ & $13.29$ &  &  &  &  & \xmark & \xmark & \xmark \\
$52.18$ & $16.48$ &  &  &  &  &  & \xmark &  \\
$54.87$ & $18.27$ &  &  &  &  & \xmark &  &  \\
$54.87$ & $18.27$ &  &  &  &  &  &  & \xmark \\
$54.87$ & $18.27$ &  &  &  &  &  &  &  \\ \bottomrule
\end{tabular}
\end{sc}
\end{small}
\end{center}
\vspace{0.25cm}
\caption{Selective ablation of NeuralCX inputs. Features that are marked \xmark\hspace{0.025cm} are replaced with noise. Ablations are sorted from top to bottom in order of disruptiveness, with the bottom row showing results from an undisturbed model. The different features are defined in Table \ref{table:features}.}
\label{table:ablation}
\vspace{-0.5cm}
\end{table}

The Embedding Model performed between Hard Negative Mining and the Distance Baseline. Interestingly, the value of $\lambda$ that maximized performance was 1.0, meaning that integrating the overt probability of $A$ under the VQA model only hurt accuracy. We observed a smooth increase in performance as we varied $\lambda$ between 0 and 1. Clearly, there is some signal in the relative position of the candidate answer embeddings around the ground truth answer, but not enough to improve on the information captured in the visual feature distance.

The NeuralCX model significantly outperformed both the Distance Baseline and the two-headed model from \citet{VQA2}. To quantify the impact of the VQA model on the performance of NeuralCX, we tested three conditions for the underlying VQA model: untrained, pretrained, and trainable. In the untrained condition, we initialized NeuralCX with an untrained VQA model. In the pretrained condition, we initialized NeuralCX with a pretrained VQA model, which was frozen during VQA-CX training. In the trainable condition, we allowed gradients generated by the loss layer of NeuralCX to backpropagate through the VQA model. We found that fine-tuning the VQA model in this manner produced small gains over the pretrained model. Meanwhile, with an untrained VQA model, the recall@5 of NeuralCX was only 2.39 points lower than with a trained model.

In the NeuralCX ablation experiments, we found that visual features were crucial to strong performance. Without any visual features, recall fell below the Distance Baseline. Both $V$ and the rank embedding appear to be especially important to the task. Intriguingly, these features also appear to be interdependent; ablating either $V$ or the rank embedding was almost as disruptive as ablating both. Meanwhile, we found that ablating the non-visual features produced a much smaller impact. While ablating $A$ resulted in a small performance drop, ablating $Q$ and $Z$ did not affect performance at all.

\section{Discussion}

\begin{figure*}[hbtp]
\begin{center}
\centerline{\includegraphics[width=1.75\columnwidth]{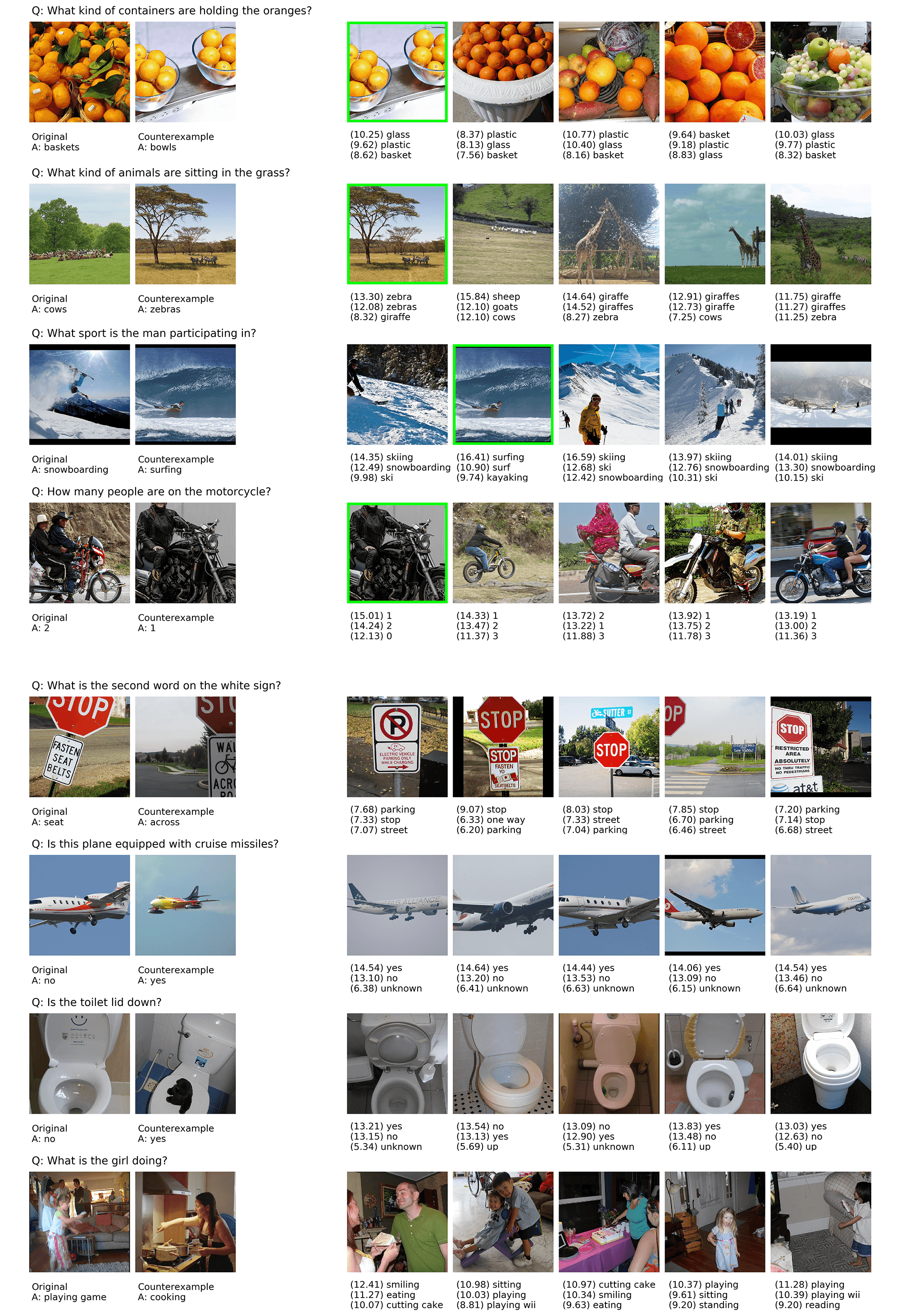}}
\caption{Qualitative results on the counterexample prediction task. Left: The original image and ground truth counterexample from VQA 2.0, along with the question and ground truth answers. Right: the top 5 counterexamples selected by NeuralCX, with the top 3 answers (as scored by a VQA model) underneath. In the top 4 rows, the model correctly identifies the correct counterexample (green outline). In the bottom 4 rows, the model fails. See the Discussion for a review of common failure modes.}
\label{fig:qualitative}
\end{center}
\end{figure*}

Our results highlight both promises and challenges associated with VQA-CX. On the one hand, the fact that NeuralCX outperforms the methods from \citet{VQA2} demonstrates that the data contain enough signal to support supervised learning. This result is especially important in light of the pronounced skew in the distribution over the rank of $I^*$ in the dataset, which makes approaches based merely on image distance unreasonably dominant. Given that the supervised neural model from \citet{VQA2} barely surpasses the Distance Baseline, it seems likely that this model overfits to the $I^*$ rank distribution. Indeed, the $K \times K$ fully-connected layer of this model inherently limits the information that can pass through to the output. Due to this bottleneck, it is unlikely that this network learns anything other than the optimal activation biases of the $K$ output units.

In contrast, we observed that NeuralCX effectively leverages both visual and semantic information. When provided with only visual features, the recall@5 for NeuralCX was 7.78 percentage points lower than when the model was provided with both visual and semantic features (Table \ref{table:ablation}). In particular, the answer embedding provides information about the semantic similarity between $A'$ and $A$, which we hypothesize allows the model to select counterexamples that are semantically distinct from the original example. The strong performance of the Embedding Model---which bases its predictions solely on answer similarity, and does not model image distance---also supports this hypothesis. Thus, while visual similarity remains a crucial feature for VQA-CX, our findings demonstrate that in order to achieve peak performance on this task, a model must also leverage semantic information.

While our results indicate that the answer embeddings encode task-relevant information, the same cannot be said for the multimodal embeddings $Z$ produced by the VQA model. In our ablation experiments, we found that replacing $Z$ and $Z'$ with noise did not affect the performance of NeuralCX. Since $Z$ is, by definition, a joint embedding of $Q$ and $V$, it is possible that $Z$ encodes redundant information. However, if this were the case, we would expect $Z$ to help the model in cases where visual features are not available. Instead, we see a significant drop in accuracy when we ablate the visual features but leave $Z$, suggesting that $Z$ does not support the recovery of visual features.

Our experiments with untrained VQA models suggest that the representations learned by the VQA model do not contain useful information for identifying counterexamples. Replacing the pretrained VQA model with an untrained version results in a decrease of only 2.39 recall@5. (Based on our ablation experiments, this performance hit is not due to the loss of $Z$, but rather, the loss of the distribution over the counterexample answer $P(\mathcal{A'})$, which is used to weight the embedding representation of $A'$). One could argue that it is unfair to expect the VQA model to provide useful information for VQA-CX, since it was not trained on this task. However, when we co-train the VQA model with NeuralCX, we find only a small performance improvement compared to the pretrained model. This result holds regardless of whether the VQA model is initialized from pretrained weights when trained on VQA-CX.

This transfer failure raises questions about the extent to which models that perform well on the VQA dataset actually learn semantic distinctions between visually-similar images. In our qualitative analysis, we found that while the VQA model often produces the correct answer, it also assigns high probability to semantically-opposite answers. For instance, when answering ``yes,'' the model's other top guesses are almost always ``no'' and ``unsure.'' Similarly, counting questions, the VQA model often hedges by guessing a range of numbers; e.g., ``1, 2, 3'' (see Fig. \ref{fig:qualitative}). While this strategy may be optimal for the VQA task, it suggests that the VQA model is effectively memorizing what types of answers are likely to result from questions and images. In other words, it is unclear from these results whether the VQA model can actually distinguish between the correct answer and other answers with opposite meanings.

While our results expose issues with existing approaches to VQA, it is important to consider two external failure modes that also affect performance on VQA-CX. First, in some cases, NeuralCX fails to fully utilize information from the VQA model. On certain examples, even when the VQA model correctly identified a particular $I'$ as producing the same answer as the original, NeuralCX still chose $I'$ as the counterexample. In other cases, NeuralCX incorrectly assigned high scores to images for which $A' \approx A$; e.g., an image of children ``playing'' was selected as a counterexample to an image of children ``playing game.'' These failures indicate that NeuralCX does not optimally leverage the semantic information provided by the VQA model.

The second failure mode arises from issues with the data itself. While the complementary pairs data in VQA 2.0 makes it possible to formalize counterexample prediction as its own machine learning task, several idiosyncrasies in the data make VQA-CX a partially ill-posed problem.

\begin{itemize}
\item There may be multiple images in $I_\mathrm{NN}$ that could plausibly serve as counterexamples. This is particularly evident for questions that involve counting (e.g., for $Q=$ ``How many windows does the house have?'' the majority of images in $I_\mathrm{NN}$ are likely to contain a different number of windows than the original image.) In many cases, our models identified valid counterexamples that were scored as incorrect, since only a single $I^* \in I_\mathrm{NN}$ is labeled as the ground truth.

\item For approximately 9\% of the examples, the counterexample answer $A^*$ is the same as $A$. This irregularity is due to the fact that the tasks of identifying counterexamples and assigning answer labels were assigned to different groups of crowd workers \citep{VQA2}. In addition to potential inter-group disagreement, the later group had no way of knowing the intentions of the former. This discontinuity resulted in a subset of degenerate ground truth counterexamples.

\item The distribution over the rank of $I^*$ within $I_\mathrm{NN}$ is not uniform; there is a strong bias towards closer nearest neighbors. In the training set, $I^*$ falls within the top 5 nearest neighbors roughly 44\% of the time.

\item Certain questions require common knowledge that VQA models are unlikely to possess (e.g., ``Is this a common animal to ride in the US?''; ``Does this vehicle require gasoline?'').

\item Other questions require specialized visual reasoning skills that, while within reach for current machine learning methods, are unlikely to be learned by general VQA architectures (e.g., ``What is the second word on the sign?'' or ``What time is on the clock?'')

\item Finally, a small portion of the questions in VQA 2.0 simply do not admit to the counterexample task. For instance, given the question, ``Do zebras have horses as ancestors?'' it is impossible to select an image, zebra or otherwise, that reverses biological fact.
\end{itemize}

While these idiosyncrasies in the data make VQA 2.0 a less than ideal domain for this task, we nevertheless view work on VQA-CX as crucial to the broader goals of representation learning. As leaderboard-based competitions like the VQA Challenge\footnote{http://www.visualqa.org/challenge.html} continue to steer research efforts toward singular objectives, we feel that auxiliary VQA tasks like counterexample prediction offer important means for cross-validating progress.

\section{Conclusion}

In this work, we explored VQA-CX: a reformulation of the VQA task that requires models to identify counterexamples. We introduced two architectures for evaluating the performance of existing VQA models on this task. Using these models, we established a new state-of-the-art for counterexample prediction on VQA 2.0. While we used a top-performing VQA model in our experiments, we found that the representations learned by this model did not contribute to performance on VQA-CX. Our results suggest that VQA models that perform well on the original task do not necessarily learn conceptual distinctions between visually-similar images. These findings raise important questions about the effectiveness of VQA in general, and the VQA 2.0 dataset in particular, as a benchmark of multimodal reasoning.

These issues occur amidst general concerns about AI interpretability. As machine learning systems assume increasing levels of decision-making responsibility in society, it is imperative that they be able to provide human-interpretable explanations \citep{doshi2017towards}. Within the domain of VQA, explanation by counterexample serves as a potentially-useful way for machines to build trust among their users \citep{VQA2}. However, this explanation modality will only serve its intended purpose if the underlying systems can meaningfully represent and distinguish between the semantic concepts encoded in images. We hope that this work will motivate the development of new benchmarks that adopt a more nuanced approach to the evaluation of visual-semantic reasoning.

\bibliographystyle{ACM-Reference-Format}
\bibliography{main}

\end{document}